\documentclass[letterpaper]{article} 
\usepackage{aaai2026}  
\usepackage{times}  
\usepackage{helvet}  
\usepackage{courier}  
\usepackage[hyphens]{url}  
\usepackage{graphicx} 
\usepackage{natbib}  
\usepackage{caption} 
\usepackage{algorithm}
\usepackage{algorithmic}
\usepackage[switch]{lineno}
\usepackage{booktabs} 
\usepackage{multirow}
\usepackage{subcaption}
\usepackage{amsmath}

\usepackage{pifont}
\usepackage{booktabs}

\newcommand{\cmark}{\textcolor{green}{\ding{51}}} 
\newcommand{\xmark}{\textcolor{red}{\ding{55}}} 

\usepackage{newfloat}
\usepackage{listings}
\DeclareCaptionStyle{ruled}{labelfont=normalfont,labelsep=colon,strut=off} 
\floatstyle{ruled}
\newfloat{listing}{tb}{lst}{}
\floatname{listing}{Listing}
\title{Benchmarking Vision-Language Models on Chinese Ancient Documents: From OCR to Knowledge Reasoning}
\author{
Haiyang Yu\textsuperscript{\rm 1}\textsuperscript{\rm 2}* , Yuchuan Wu\textsuperscript{\rm 1}*, Fan Shi\textsuperscript{\rm 1}*, Lei Liao\textsuperscript{\rm 2}*, Jinghui Lu\textsuperscript{\rm 2}, \\ Xiaodong Ge\textsuperscript{\rm 1}, Han Wang\textsuperscript{\rm 2}, Minghan Zhuo\textsuperscript{\rm 1}, Xuecheng Wu\textsuperscript{\rm 2}, Xiang Fei\textsuperscript{\rm 2}, Hao Feng\textsuperscript{\rm 2}, Guozhi Tang\textsuperscript{\rm 2}, An-Lan Wang\textsuperscript{\rm 2}, Hanshen Zhu\textsuperscript{\rm 2}, Yangfan He\textsuperscript{\rm 2}, Quanhuan Liang\textsuperscript{\rm 2}, Liyuan Meng\textsuperscript{\rm 2}, Chao Feng\textsuperscript{\rm 2},\\ Can Huang\textsuperscript{\rm 2}\dag, Jingqun Tang\textsuperscript{\rm 2}\dag \ddag, Bin Li\textsuperscript{\rm 1}\dag
}
\affiliations{
    \textsuperscript{\rm 1} Fudan University\\
    \textsuperscript{\rm 2} ByteDance Inc.\\
   
}

\usepackage{bibentry}

\begin{document}
\nocopyright
\maketitle
\begingroup
\renewcommand\thefootnote{}%
\footnotetext{* Equal contribution, \dag\ Corresponding author, \ddag \ Project lead.}%
\endgroup

\begin{abstract}
Chinese ancient documents, invaluable carriers of millennia of Chinese history and culture, hold rich knowledge across diverse fields but face challenges in digitization and understanding—traditional methods only scan images, while current Vision-Language Models (VLMs) struggle with their visual/linguistic complexity. Existing document benchmarks focus on English printed texts or simplified Chinese, leaving a gap for evaluating VLMs on ancient Chinese documents. To address this, we present AncientDoc, the first benchmark for Chinese ancient documents, designed to assess VLMs from OCR to knowledge reasoning. AncientDoc includes five tasks (page-level OCR, vernacular translation, reasoning-based QA, knowledge-based QA, linguistic variant QA) and covers 14 document types, over 100 books, and about 3,000 pages. Based on AncientDoc, we evaluate mainstream VLMs using multiple metrics, supplemented by a human-aligned large language model for scoring. The benchmark are available at \url{https://bytedance.github.io/AncientDoc}.

\end{abstract}

\section{Introduction}

Chinese ancient documents, which carry thousands of years of Chinese history and culture, are treasure troves of knowledge spanning history, philosophy, medicine, astronomy, etc. They are invaluable cultural heritage for both China and the world. With the wave of digitization of Chinese ancient documents in libraries and museums, many captured Chinese ancient document images are produced. However, traditional digitization methods only stay at the level of image scanning, while many downstream applications (knowledge mining, historical exploration) urgently need the ability to deeply understand the content of ancient documents. At the same time, parsing and understanding Chinese ancient document images pose huge challenges, including visual complexity, linguistic complexity, and poor adaptability of current vision-language models (VLMs).

Existing general document understanding datasets and benchmarks (\textit{e.g.}, DocVQA~\cite{mathew2021docvqa}) are mainly based on printed documents and are predominantly in English. Even Chinese-related datasets (\textit{e.g.}, CN-DocVQA) only involve simplified Chinese characters, which is totally different from Chinese ancient documents. In addition, with the development of large vision-language models (VLMs), an increasing number of VLMs have acquired the capabilities of document OCR and understanding. However, for VLMs, there is currently no benchmark that can systematically evaluate their OCR and understanding capabilities on Chinese ancient documents.

To address these issues, we construct the first Chinese ancient document benchmark called \textit{AncientDoc}, which is used to comprehensively evaluate the capabilities of VLMs ranging from OCR to knowledge reasoning. AncientDoc includes five tasks: page-level OCR, vernacular translation, reasoning-based QA (question answering), knowledge-based QA, and linguistic variant QA. AncientDoc has diverse sources, covering 14 types of ancient documents (such as collected works and Chuci-style poetry), approximately 100 books, and a total of 3000 document pages. In addition, we have adopted several evaluation metrics to evaluate most mainstream VLMs on the five tasks. To supplement these metrics, we additionally utilize a large language model to score the predictions of VLMs.

\begin{table*}[ht]
\centering
\scalebox{0.9}{
\begin{tabular}{lccccccc}
\toprule
\textbf{Task}  & \textbf{DocVQA} & \textbf{TKH} & \textbf{MTH} & \textbf{OCRBench} & \textbf{OCRBench v2} & \textbf{AncientDoc} \\
\midrule
Page-level OCR        & \xmark & \cmark & \cmark & \cmark & \cmark & \cmark \\
Vernacular Translation            & \xmark & \xmark & \xmark & \xmark & \xmark & \cmark \\
Reasoning-based QA  & \cmark & \xmark & \xmark & \cmark & \cmark & \cmark \\
Knowledge-based QA         & \xmark & \xmark & \xmark & \xmark & \xmark & \cmark \\
Linguistic Variant QA     & \xmark & \xmark & \xmark & \xmark & \xmark & \cmark \\
\bottomrule
\end{tabular}}
\caption{Comparison of task type between different bemchmarks.}
\label{task_cmp}
\end{table*}

In summary, the contributions of our work are as follows:
\begin{itemize}
    \item We propose the first benchmark (AncientDoc) for Chinese ancient documents, aiming to comprehensively evaluate existing vision-language models from OCR to knowledge reasoning.
    \item AncientDoc contains five tasks for evaluating VLMs: page-level OCR, vernacular translation, reasoning-based QA, knowledge-based QA, and linguistic variant QA. It covers 14 types of ancient documents, with 3,000 page images extracted from over 100 ancient books.
    \item We have conducted a comprehensive evaluation of existing mainstream vision-language models with various metrics. In addition, we also adopt a large language model that is the most consistent with human scoring to evaluate them.
\end{itemize}

\section{Related Work}

In recent years, document understanding tasks~\cite{borchmann2021due,ma2024mmlongbench,tanaka2024instructdoc} have continuously expanded from traditional Optical Character Recognition (OCR)~\cite{kang2022pay,yin2017scene,ingle2019scalable} to higher-level semantic understanding tasks~\cite{ding2023vqa,zhang2024cream}, such as question answering~\cite{mishra2019ocr,ding2024mvqa,kang2024multi}, translation~\cite{zhang2018improving,wang2023document}, and structured information extraction~\cite{jaume2019funsd,huang2022layoutlmv3}. With the increasing complexity of tasks, multimodal document datasets have become increasingly abundant, providing an important basis for evaluating the capabilities of different models. However, most existing datasets still suffer from limitations in task dimensions, making it difficult to fully cover the multi-level cognitive and language transfer tasks involved in complex Chinese ancient documents.

Among existing datasets, DocVQA~\cite{mathew2021docvqa} is an early representative multimodal dataset focusing on visual document question answering tasks, emphasizing linguistic reasoning on structured and unstructured text in document images. The question format of DocVQA usually relies on OCR results rather than the linguistic content, and its core task is text logical reasoning, which is suitable for evaluating the text understanding and document reading abilities of models. However, this dataset does not include page-level OCR task, so it cannot examine the recognition robustness of models when faced with complex visual inputs. Therefore, it is still insufficient in evaluating the comprehensive capabilities of models.

Additionally, TKH~\cite{yang2018dense} and MTH~\cite{yang2018dense} focus more on character-level recognition of Chinese historical documents, mainly used to evaluate the OCR capabilities of models on ancient books with low quality and numerous variant characters. Both are derived from real historical materials and have strong characteristics of ancient book images, such as vertical typesetting and cursive script, making them suitable for basic OCR evaluation datasets. However, neither of them involves tasks at the level of language understanding or language generation, so they cannot be used to evaluate the model's ability to understand the semantics, grammatical structure, or background knowledge of ancient Chinese. They also lack vernacular output or question-answer interaction forms, which limits their value in language transfer ability analysis.

To alleviate the above problem of single tasks, OCRBench~\cite{liu2024ocrbench} and its enhanced version OCRBench v2~\cite{fu2024ocrbench} have introduced more diverse settings, covering multiple subtasks, including character recognition, text localization, and some document-level question answering and logical reasoning problems. Among them, OCRBench v2 has improved in the task complexity compared to the first version, making it suitable for evaluating the comprehensive performance of multimodal models in real scenarios. However, these two datasets mainly focus on modern Chinese or English documents. At the same time, they do not cover tasks such as translation, knowledge question answering, or language style transfer, so they are difficult to use for evaluating the generalization performance of models in cross-lingual and cross-style understanding abilities.


To this end, we have constructed a multi-task evaluation benchmark dataset \textit{AncientDoc} for Chinese ancient documents. This dataset not only includes traditional OCR recognition tasks but also systematically integrates multiple high-level language understanding tasks. The comparison of task diversity is shown in Tab.~\ref{task_cmp}.

\section{Benchmark Construction}

\begin{figure*}[t]
\centering
\includegraphics[width=1.0\textwidth]{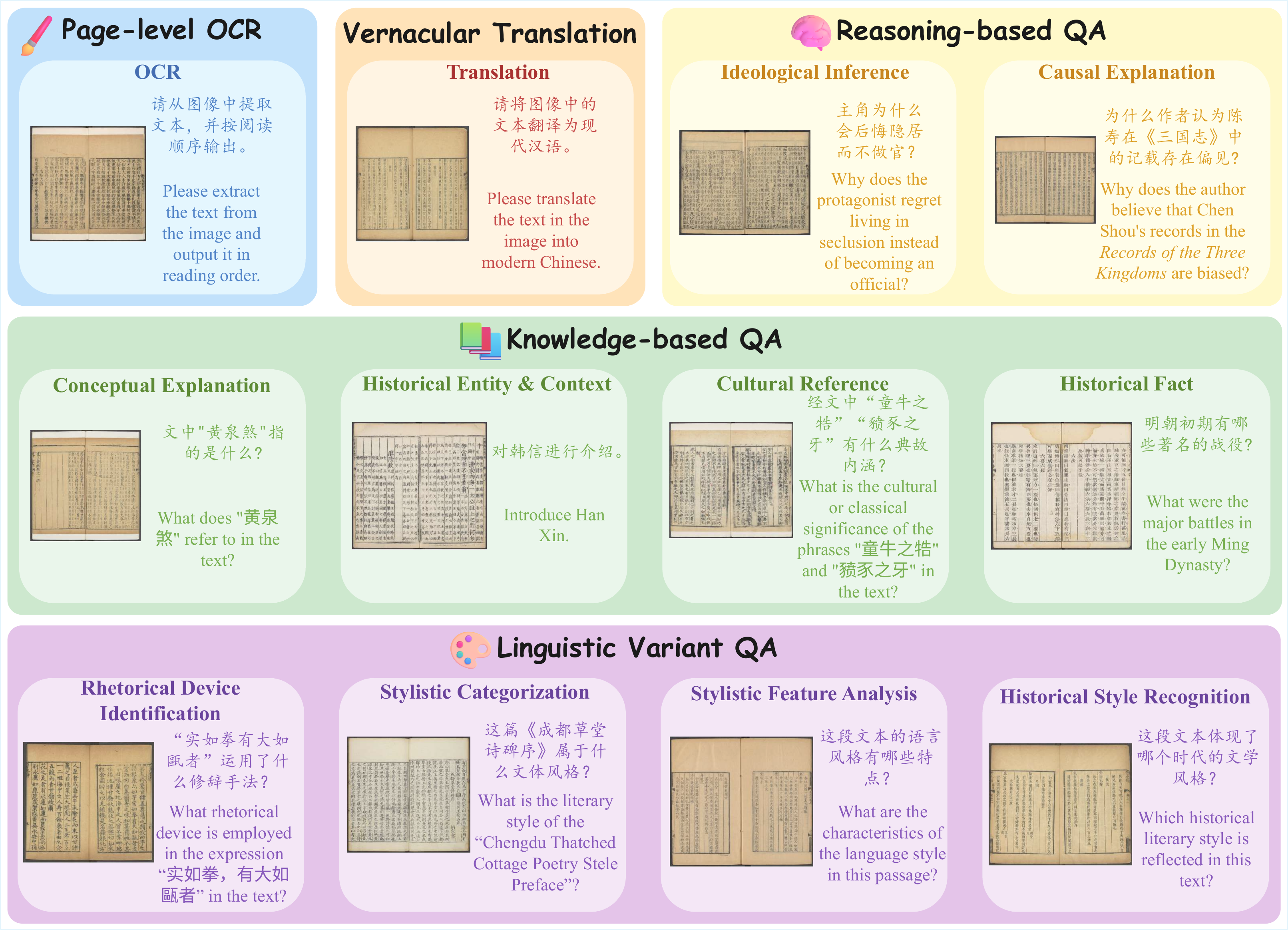} 
\caption{Some examples of each task in AncientDoc.}
\label{sample}
\end{figure*}

\subsection{Task Definition}

In this paper, we define five tasks for evaluating VLMs on Chinese ancient documents: page-level OCR, vernacular translation, reasoning-based QA (question answering), knowledge-based QA, and linguistic variant QA. Some examples of these tasks are shown in Fig.~\ref{sample}. 

\textbf{Page-level OCR}: This task aims to directly extract complete and correctly ordered text content from an entire page of ancient documents, without relying on the character detection~\cite{tian2016detecting,zhou2017east}, segmentation~\cite{he2016reading,xu2025improving}, and recognition~\cite{akoushideh2025persian,cui2025enhanced} processes in traditional OCR systems. This task presents the following challenges: 1) \textit{Vertical texts}: Most Chinese ancient documents are vertically typeset from right to left, so the model needs to understand the correct reading order and the rules for line breaks within columns. 2) \textit{Various annotations}: Chinese ancient documents sometimes contain interlinear notes, comments, small characters, postscripts, etc., requiring the model to have visual and semantic filtering capabilities. 3) \textit{Traditional Chinese characters}: There are a large number of traditional Chinese characters or obsolete glyphs in Chinese ancient documents. Nevertheless, training samples of these characters are scarce.

\textbf{Vernacular Translation}: It aims to translate the Chinese texts in ancient documents into modern common vernacular expressions, enabling non-professional readers to understand the meaning of original texts and providing a clearer linguistic foundation for downstream tasks (such as question answering and summarization.). Unlike translation between languages, this task is intralingual translation. The difficulties of this task are: 1) \textit{Polysemy}: Some ancient Chinese words often have multiple meanings, and the model needs to understand the context or even the semantics of the entire paragraph to select the correct interpretation. 2) \textit{Semantic punctuation}: There is a lack of punctuation in the text of Chinese ancient documents. Therefore, the model needs to insert punctuation based on semantic understanding.

\textbf{Reasoning-based QA}: Based on an image of a page from ancient documents, reasoning-based QA aims at extracting implicit information to answer questions that are not directly stated. Different from extractive question answering, reasoning-based QA requires the model to have the ability to understand and derive deep-level information such as facts, causality, and semantic relationships. The reasoning-based QA task is one of the advanced types of OCR-free document understanding. It integrates the understanding of graphic information, the interpretation of ancient Chinese language, and the connection and deduction of knowledge and logic, serving as an important scenario to test the deep language understanding and multi-step reasoning abilities of large models.

\textbf{Knowledge-based QA}: This task requires the model to answer questions related to objective knowledge in ancient documents, including time, place names, objects, medical terms, etc. Although this task belongs to factual QA, it still differs from encyclopedia QA. This task requires the model to understand the knowledge expression methods in ancient languages and have a certain reserve of historical and cultural background knowledge. Knowledge-based QA not only tests the model's ability to handle the explicit expression of knowledge but also deeply challenges its ability to summarize and infer knowledge under vague descriptions, putting forward new requirements for OCR-free document understanding and the knowledge transfer ability of VLMs in historical corpora.

\textbf{Linguistic Variant QA}: It aims to evaluate the model's ability to understand and reason about variant phenomena in ancient Chinese, such as language styles, rhetorical methods, and stylistic features. This task requires the model to generate or answer relevant questions around the aforementioned linguistic features. The linguistic variant question answering task is a key task connecting the understanding of linguistic artistry and generative language expression, representing a high-level language ability that moves from text recognition to the mastery of stylistic features. This task not only assesses whether the model "understands ancient Chinese" but also evaluates whether it understands how ancient Chinese is written, what its style is, and how rhetoric affects semantics.

\subsection{Data Curation}

\begin{figure}[t]
    \centering
    \begin{subfigure}[b]{0.2\textwidth} 
        \centering
        \includegraphics[width=\textwidth]{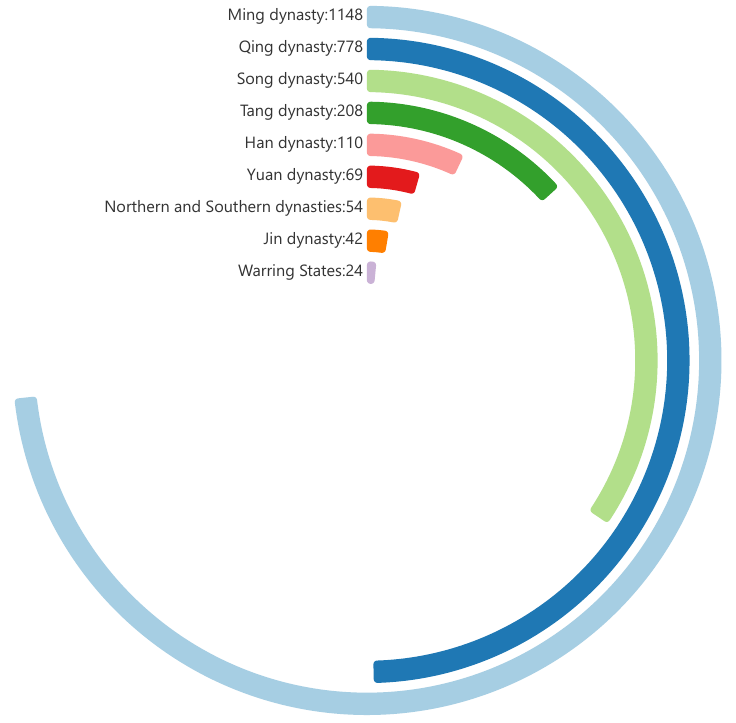} 
        \caption{}
        \label{fig:sub1}
    \end{subfigure}
    \hfill 
    \begin{subfigure}[b]{0.25\textwidth} 
        \centering
        \includegraphics[width=\textwidth]{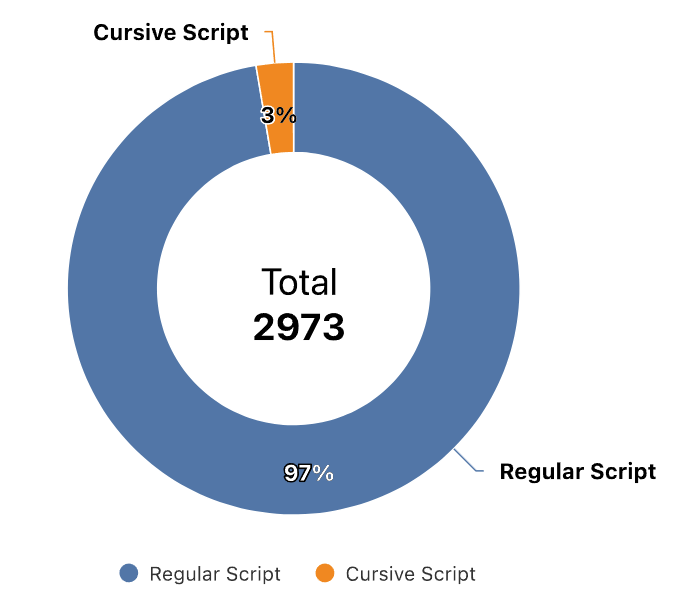}
        \caption{}
        \label{fig:sub2}
    \end{subfigure}
    
    \caption{(a) The distribution of page counts across different dynasties in AncientDoc. (b) The proportion of cursive script and regular script in AncientDoc.}
    \label{fig:double}
\end{figure}

\subsubsection{Data Resources}
The original images of the proposed AncientDoc are mainly derived from the digitized resources of Chinese ancient documents held by the Harvard Library\footnote{https://hollis.harvard.edu/}. The digitized collection of ancient books in this library covers multiple dynasties and fields, featuring high scanning quality and rich typesetting styles, which provides a solid foundation for constructing a multi-task dataset of Chinese ancient documents. In combination with the requirements of semantic-related tasks, all collected ancient documents have undergone manual verification and classification. Finally, we divide the ancient documents into 14 semantic categories, including ``collected works'',``Chuci-style Poetry'',``Literary Criticism of Poetry and Prose'', ``Eclectics'', etc. All categories and corresponding explanations will be detailed in the supplementary materials.

\subsubsection{Data Collection}
In the process of data collection, we select representative Chinese printed ancient documents dating from the Qing Dynasty and earlier (The distribution of page counts across different dynasties is shown in Fig.~\ref{fig:double}(a)). To ensure the usability and challenge of the collected documents in terms of visual quality, linguistic content, and task adaptability, we establish the following priority criteria for collection:
\begin{itemize}
    \item \textit{Vertical typesetting with traditional Chinese characters}: The selected pages conform to the typographic style of traditional ancient documents and feature structurally challenging information. Thus, they are suitable for evaluating the ability of OCR-free models to understand reading directions and inter-column structures.
    \item \textit{Clear fonts with partial degradation or damage}: They should cover real-world scenarios such as ink blurring, simulating common degradation in the digitization of ancient documents. These selected samples can evaluate the robustness of models against low-quality inputs.
    \item \textit{Content with potential for linguistic, structural, knowledge-based, and reasoning tasks}: Poems, annotations, encyclopedias, historical biographies, and medical theories are widely chosen since they have high semantic density and are suitable for setting multi-level understanding tasks (\textit{e.g.}, vernacular translation, reasoning-based QA, knowledge-based QA);
    \item \textit{High readability of images for convenient annotation}: We select versions with clear scans and complete page numbers to ensure effective OCR annotations, translation alignment, etc.

\end{itemize}

Ultimately, we collect approximately 100 ancient books for AncientDoc, covering various themes and styles and containing a total of about 3,000 pages of image data.

\subsubsection{Data Annotation}

To efficiently construct a high-quality multi-task dataset for Chinese ancient documents, we adopt an annotation strategy combining pre-annotation by large language models and manual proofreading. The specific process is as follows.

First, we use Qwen2.5-VL-72B~\cite{bai2025qwen2} model to formulate questions and corresponding answers that align with the requirements of different defined tasks and the inherent characteristics of each category of ancient documents. For each collected ancient document page, we construct one QA pair for page-level OCR and vernacular translation, and two QA pairs for the remaining three tasks. In the following, we conduct comprehensive manual review and revision of the generated results, which means that some pre-constructed QA pairs by Qwen may be filtered. The manual revisions cover the following aspects: OCR structure correction, translation quality improvement and consistency check across multi-tasks. Details of them are described in the supplementary materials.

\subsection{Statistical Analysis}

\begin{figure}[t]
\centering
\includegraphics[width=0.45\textwidth]{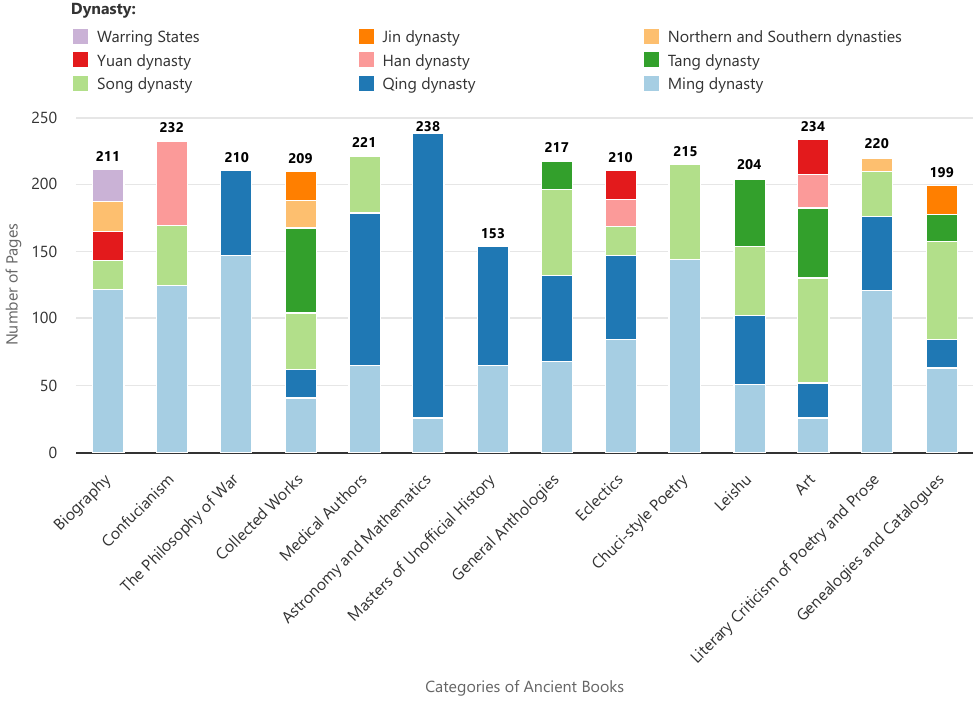} 
\caption{The page count distribution of different categories of ancient books.}
\label{page_count}
\end{figure}

\begin{table*}[htbp]
\centering

\scalebox{0.9}{
\begin{tabular}{lcccccc}
\toprule  
& Pearson & Spearman & Kendall & MSE & MAE & Bias \\
\midrule  
Qwen2.5-VL-72B~\cite{bai2025qwen2} & \underline{0.833} & \underline{0.824} & \textbf{0.718} & \underline{3.415} & \underline{1.364} & 1.044 \\
Gemini~\cite{comanici2025gemini} & 0.829 & 0.822 & 0.67 & 4.463 & 1.644 & -1.436 \\
Doubao~\cite{seed2025seed1_5vl} & 0.695 & 0.774 & 0.637 & 10.739 & 2.696 & -2.616 \\
Qwen-Plus~\cite{qwen2024qwen25} & 0.767 & 0.743 & 0.618 & 3.631 & 1.54 & \textbf{0.724} \\
GPT-4o~\cite{hurst2024gpt} & \textbf{0.846} & \textbf{0.837} & \underline{0.689} & \textbf{2.939} & \textbf{1.32} & \underline{-0.896} \\
\bottomrule  
\end{tabular}}
\caption{Comparison of differences between large model scoring and human scoring.}
\label{score_cmp}
\end{table*}

\begin{table*}[t]
  \centering
  \scalebox{0.80}{
  \begin{tabular}{clccccc}
    \toprule
    & \textbf{Model} & \textbf{CER} & \textbf{Char Precision} & \textbf{Char Recall} & \textbf{Char F1} & \textbf{GPT-4o} \\
    \midrule
    \multirow{16}{*}{\parbox{2cm}{ Open-source MLLMs}} & \multicolumn{5}{c}{\textbf{2B-4B MLLMs}} \\
    \cmidrule(lr){2-7}
    & InternVL2.5-2B~\shortcite{chen2024expanding} & 120.72 & 1.76 & 1.69 & 1.72 & 2.03\\
    & InternVL3-2B~\shortcite{zhu2025internvl3} & 95.59 & 2.46 & 1.79 & 2.07 & 2.52\\
    & Qwen2.5-VL-3B~\shortcite{bai2025qwen2} & 50.36 & 9.44 & 9.02 & 9.23 & 5.56\\
    & Qwen2.5-Omni-3B~\shortcite{xu2025qwen2} & 63.42 & 5.4 & 4.78 & 5.07 & 4.56\\
    & InternVL2.5-4B~\shortcite{chen2024expanding} & 79.75 & 5.27 & 4.05 & 4.58 & 3.65\\
    \addlinespace
    \cmidrule(lr){2-7}
    & \multicolumn{5}{c}{\textbf{7B-8B MLLMs}} \\
    \cmidrule(lr){2-7}
    & Qwen2.5-VL-7B~\shortcite{bai2025qwen2} & \underline{35.47} & 12.95 & 12.75 & 12.85 & \textbf{6.37}\\
    & Qwen2.5-Omni-7B~\shortcite{xu2025qwen2} & 70.46 & 3.44 & 3.35 & 3.39 & 4.30\\
    & LLaVA-1.5-7B~\shortcite{liu2024improved} & 129.71 & 0.04 & 0.03 & 0 & 0.05\\
    & InternVL2.5-8B~\shortcite{chen2024expanding} & 96.88 & 2.87 & 2.59 & 2.72 & 3.00\\
    & InternVL3-8B~\shortcite{zhu2025internvl3} & 51.8 & 7.3 & 6.67 & 6.97 & 5.06\\
    \addlinespace
    \cmidrule(lr){2-7}
    & \multicolumn{5}{c}{\textbf{70B+ MLLMs}} \\
    \cmidrule(lr){2-7}
    & Qwen2.5-VL-72B~\shortcite{bai2025qwen2} & 58.83 & 10.62 & 9.56 & 10.06 & 5.73\\
    & LLaVA-OneVision-72B~\shortcite{li2024llava} & 190.67 & 0.29 & 0.43 & 0.25 & 0.01\\
    & InternVL3-78B~\shortcite{zhu2025internvl3} & 78.67 & 4.7 & 4.89 & 4.79 & 4.04\\

    \midrule
    \multirow{4}{*}{\parbox{2cm}{ Closed-source MLLMs}} & Doubao-V2~\cite{seed2025seed1_5vl} & 71.95 & \underline{13.14} & \textbf{20.44} & \underline{16} & \underline{6.27}\\
    & Qwen-VL-Max~\shortcite{Qwen-VL} & 66.39 & 9.62 & 9.93 & 9.77 & 5.70\\
    & Gemini2.5-Pro~\shortcite{comanici2025gemini} & \textbf{32.03} & \textbf{17.73} & \underline{18.53} & \textbf{18.12} & 6.08\\
    & GPT-4o~\shortcite{hurst2024gpt} & 75.1 & 4.83 & 2.72 & 3.48 & 2.97\\
    \bottomrule
  \end{tabular}}
  \caption{Evaluation on page-level OCR of AncientDoc.} 
  \label{ocr}
\end{table*}

We have collected a total of 2,973 pages of Chinese ancient documents and conducted a systematic analysis of them from multiple dimensions. Firstly, in terms of chronological distribution, the data covers major dynasties from the Warring States, Qin, and Han dynasties to the Ming and Qing dynasties, showing a good historical span (as shown in Fig.~\ref{fig:double}(a)). Among them, images from the Ming Dynasty (1,148 pages) and the Qing Dynasty (778 pages) are the most abundant, accounting for approximately 65\% of the total pages, indicating that the preservation and circulation of documents from the Ming and Qing dynasties are relatively complete. Next are those from the Song Dynasty (540 pages) and the Tang Dynasty (208 pages), while documents from the Han, Yuan, and Southern and Northern Dynasties are relatively scarce.

In terms of genre distribution, the ancient book samples cover 14 categories of mainstream traditional literature, including biographies, Confucianism, philosophy of war, collected works, medical authors, astronomy and mathematics, masters of unofficial history, general anthologies, eclectics, Chuci-style poetry,  Leishu, art, literary criticism of poetry and prose, and genealogies and catalogues. Statistical results (see Fig.~\ref{page_count}) show that the top three categories in terms of page count are astronomy and mathematics (238 pages), Confucianism (232 pages), and art (234 pages). 


In terms of font style, the overall writing of the data is standardized and clear. Approximately 97\% of the ancient book pages are in regular script, and only 3\% are in cursive script (see Fig.~\ref{fig:double}(b)), ensuring the recognizability for manual annotating. Overall, the proposed AncientDoc has good representativeness in three aspects: chronological span, genre diversity, and writing style.



\subsection{Evaluation Metric}
\label{Metric}

To assess the accuracy of page-level OCR, we use the following four metrics: CER (Character Error Rate), Char Precision, Char Recall and Char F1. The detailed description of them are shown in the supplementary material. Differently, when evaluating the remaining four tasks (\textit{i.e.}, vernacular translation, reasoning-based QA, knowledge-based QA and linguistic variant QA), we use CHRF++~\cite{popovic2017chrf++} and BERTScore (BS-F1) as the key metrics~\cite{zhang2019bertscore}.

Considering the potential irrationality of the aforementioned hard metrics (for example, in vernacular translation, inverted word order may lead to a low CHRF++ score), we also use large language models to score the predictions and labels of each task (on a scale of 0 to 10). We utilize Qwen2.5-72B, GPT4o, Gemini, Doubao, and Qwen-plus to score the outputs of the evaluated models against the labels. To select the model that best aligns with human judgment, we compare the scoring results of select models with human ratings. Specifically, we select 50 QA pairs from each task (250 QA pairs in total) and feed them to the five candidate models and human raters for scoring (on a 0-10 scale). Then, we calculate the differences between them based on six metrics: Pearson, Spearman, Kendall, MSE, MAE, and Bias. The details of them can be found in the supplementary materials. Through the results in Tab.~\ref{score_cmp} We observe that GPT-4o is closer to human ratings across several metrics. Therefore, we ultimately choose GPT-4o for scoring. In the main text, we only show the evaluation results of BERTScore and GPT-4o score. The results of CHRF++ are shown in the supplementary materials.

\section{Results}

\subsection{Model Selection}

For evaluation, we select models including GPT-4o~\cite{hurst2024gpt}, Qwen series~\cite{bai2025qwen2,Qwen-VL,xu2025qwen2}, InternVL series~\cite{chen2024expanding,zhu2025internvl3}, as well as Doubao~\cite{seed2025seed1_5vl}, Gemini~\cite{comanici2025gemini}, LLaVA~\cite{liu2024improved}, DeepSeek~\cite{wu2024deepseek}, etc. Considering the limited pages, we only show part of results in the main text. More evaluation results are shown in the supplementary material.

\begin{table}[t]
  \centering
  \scalebox{0.8}{
  \begin{tabular}{clccccc}
    \toprule
    & \textbf{Model} & \textbf{GPT-4o} & \textbf{BS-F1} \\
    \midrule
    \multirow{14}{*}{\parbox{2cm}{Open-source MLLMs}} & \multicolumn{3}{c}{\textbf{2B-4B MLLMs}} \\
    \cmidrule(lr){2-4}
    & InternVL2.5-2B & 0.03 & 53.92 \\
    & InternVL3-2B & 0.10 & 51.83 \\
    & Qwen2.5-VL-3B & 0.66 & 55.2 \\
    & Qwen2.5-Omni-3B & 0.56 & 56.8 \\
    & InternVL2.5-4B & 0.53 & 58.46 \\
    \addlinespace
    \cmidrule(lr){2-4}
    & \multicolumn{3}{c}{\textbf{7B-8B MLLMs}} \\
    \cmidrule(lr){2-4}
    & Qwen2.5-VL-7B & 2.30 & 65.59 \\
    & Qwen2.5-Omni-7B & 0.46 & 53.99 \\
    & LLaVA-1.5-7B & 0.01 & 50 \\
    & InternVL2.5-8B & 0.58 & 59.24 \\
    & InternVL3-8B & 0.29 & 53.16 \\
    \addlinespace
    \cmidrule(lr){2-4}
    & \multicolumn{3}{c}{\textbf{70B+ MLLMs}} \\
    \cmidrule(lr){2-4}
    & Qwen2.5-VL-72B & 2.98 & 69.87 \\
    & LLaVA-OneVision-72B & 0.03 & 48.83 \\
    & InternVL3-78B & 1.21 & 62.4 \\

    \midrule
    \multirow{4}{*}{\parbox{2cm}{Closed-source MLLMs}} & Qwen-VL-Max & \underline{3.22} & \underline{71.03} \\
    & Doubao-V2 & 2.12 & 52.39 \\
    & Gemini2.5-Pro & \textbf{4.72} & \textbf{72.5} \\
    & GPT-4o & 0.92 & 58.86 \\
    \bottomrule
  \end{tabular}}
    \caption{Evaluation on vernacular transoaltion of AncientDoc.}

  \label{translation}
\end{table}


\begin{table}[t]
  \centering
  \scalebox{0.8}{
  \begin{tabular}{clcc}
    \toprule
    & \textbf{Model} & \textbf{GPT-4o} & \textbf{BS-F1} \\
    \midrule

    \multirow{14}{*}{\parbox{2cm}{Open-source MLLMs}} & \multicolumn{3}{c}{\textbf{2B-4B MLLMs}} \\
    \cmidrule(lr){2-4}
    & InternVL2.5-2B & 3.55 & 65.29 \\
    & InternVL3-2B & 4.39 & 66.45 \\
    & Qwen2.5-VL-3B & 4.67 & 65.83 \\
    & Qwen2.5-Omni-3B & 4.90 & 66.96 \\
    & InternVL2.5-4B & 4.75 & 66.12 \\

    \addlinespace
    \cmidrule(lr){2-4}
    & \multicolumn{3}{c}{\textbf{7B-8B MLLMs}} \\
    \cmidrule(lr){2-4}
    & Qwen2.5-VL-7B & 6.44 & 69.96 \\
    & Qwen2.5-Omni-7B & 6.05 & 68.70 \\
    & LLaVA-1.5-7B & 1.64 & 61.69 \\
    & InternVL2.5-8B & 5.93 & 68.40 \\
    & InternVL3-8B & 4.98 & 65.62 \\

    \addlinespace
    \cmidrule(lr){2-4}
    & \multicolumn{3}{c}{\textbf{70B+ MLLMs}} \\
    \cmidrule(lr){2-4}
    & Qwen2.5-VL-72B & 7.43 & \textbf{71.40} \\
    &  LLaVA-OneVision-72B & 5.48 & 67.20 \\
    &  InternVL3-78B & 5.18 & 65.99 \\

    \midrule
    \multirow{4}{*}{\parbox{2cm}{Closed-source MLLMs}} & Qwen-VL-Max & \underline{7.46} & \underline{71.30} \\
    & Doubao-V2 & 7.40 & 68.78 \\
    & Gemini2.5-Pro & \textbf{7.76} & 69.33 \\
    & GPT-4o & 6.92 & 70.52 \\
    \bottomrule
  \end{tabular}}
    \caption{Evaluation on reasoning-based QA of AncientDoc.}
  \label{reasoning}

\end{table}

\subsection{Main Results}

\textbf{Results in Page-level OCR.}
The results shown in Tab.~\ref{ocr} demonstrate that Gemini 2.5-pro has the best overall performance, with a Char F1 of 18.12 and a CER of only 32.03. It achieves the best results in three metrics and the second-best in another, demonstrating strong accuracy and stability. Doubao-V2 achieves sub-optimal or better results in all metrics except CER. However, its excessively high CER indicates insufficient ability in recognizing ancient characters. Additionally, models in the Qwen2.5 series all perform well in the page-level OCR task of AncientDoc. Interestingly, Qwen2.5-VL-7B outperforms Qwen2.5-VL-72B across the board and even achieves the highest score in GPT-4o scoring, which indicates that the pure OCR task only requires models to recognize characters and output them in order, and excessive understanding and reasoning capabilities may instead interfere with the output results. Overall, the Qwen2.5 series and Gemini models have greater practical value in page-level OCR, making them the preferred options for current ancient OCR tasks.

\noindent\textbf{Results in Vernacular Translation.} From the experimental results shown in Tab.~\ref{translation}, Gemini 2.5-pro achieves the score of 72.5 and 4.72 in BERTScore and GPT-4o scoring respectively, demonstrating its excellent performance in vernacular translation tasks. It outperforms all open-source and closed-source comparison models, emerging as the best baseline for vernacular translation tasks. Among the Qwen series, Qwen-VL-max (71.03), Qwen2.5-VL-72B (69.87), and Qwen2.5-VL-7B (65.59) show significant advantages over other open-source models. This result indicates that the Qwen2.5 series possesses stable capabilities in understanding ancient texts, semantic modeling, and expression in modern Chinese. In contrast, the InternVL and LLaVA series models generally score lower in this task, suggesting that their multimodal modeling focuses more on general tasks, with a clear lack of specialized processing for ancient Chinese corpora. Additionally, it is noteworthy that the performance of the closed-source model GPT-4o (BERTScore 58.86) even lags behind several open-source models with 7B parameters, such as Qwen2.5-VL-7B. Last but not least, due to the diversity of vernacular translation results, GPT-4o's scores are generally low. With a maximum score of 10, the average score of the top-performing model only reaches 4.72.

\noindent\textbf{Results in Reasoning-based QA.} From the results in Tab.~\ref{reasoning}, Qwen2.5-VL-72B achieves the highest BERTScore by a significant margin, indicating its superior performance in understanding the context of ancient texts and the implicit reasoning capability. Notably, despite having a much smaller parameter size than large models with 70B+ parameters, Qwen2.5-VL-7B still maintains a near-top BERTScore and GPT-4o scores, with only a small gap from models like GPT-4o and Qwen2.5-VL-72B. This demonstrates the model's excellent reasoning capability under low-parameter conditions. From 3B to 72B, the models' BERTScore exhibits an overall upward trend, reflecting a significant positive correlation between model size and semantic reasoning ability. Consistent with the vernacular translation task, the InternVL and LLaVA series models also achieve relatively low BERTScore in reasoning-based QA, indicating their difficulty in establishing effective contextual understanding and causal reasoning.


\begin{table}[t]
  \centering
  \scalebox{0.8}{
  \begin{tabular}{clcc}
    \toprule
    & \textbf{Model} & \textbf{GPT-4o} & \textbf{BS-F1} \\
    \midrule

    \multirow{14}{*}{\parbox{2cm}{Open-source MLLMs}} & \multicolumn{3}{c}{\textbf{2B-4B MLLMs}} \\
    \cmidrule(lr){2-4}
    & InternVL2.5-2B & 2.48 & 64.24 \\
    & InternVL3-2B & 3.52 & 66.21 \\
    & Qwen2.5-VL-3B & 3.77 & 62.86 \\
    & Qwen2.5-Omni-3B & 4.11 & 66.10 \\
    & InternVL2.5-4B & 3.92 & 64.81 \\

    \addlinespace
    \cmidrule(lr){2-4}
    & \multicolumn{3}{c}{\textbf{7B-8B MLLMs}} \\
    \cmidrule(lr){2-4}
    & Qwen2.5-VL-7B & 5.23 & 66.75 \\
    & Qwen2.5-Omni-7B & 4.94 & 66.85 \\
    & LLaVA-1.5-7B & 0.78 & 60.37 \\
    & InternVL2.5-8B & 4.88 & 67.68 \\
    & InternVL3-8B & 4.31 & 63.60 \\

    \addlinespace
    \cmidrule(lr){2-4}
    & \multicolumn{3}{c}{\textbf{70B+ MLLMs}} \\
    \cmidrule(lr){2-4}
    & Qwen2.5-VL-72B & \underline{6.84} & \underline{69.15} \\
    & LLaVA-OneVision-72B & 5.04 & 66.31 \\
    & InternVL3-78B & 5.18 & 65.79 \\

    \midrule
    \multirow{4}{*}{\parbox{2cm}{Closed-source MLLMs}} & Qwen-VL-Max & 6.78 & 68.67 \\
    & Doubao-V2 & \textbf{7.36} & \underline{69.15} \\
    & Gemini2.5-Pro & \textbf{7.36} & 68.94 \\
    & GPT-4o & 6.53 & \textbf{70.01} \\
    \bottomrule
  \end{tabular}}
  
    \caption{Evaluation on knowledge-based QA of AncientDoc.}
\label{knowledge}
\end{table}

\noindent\textbf{Results in Knowledge-based QA.} From the experimental results in Tab.~\ref{knowledge}, GPT-4o ranks first with an excellent BERTScore of 70.01, indicating its strong language modeling capabilities and extensive ancient Chinese knowledge. In the metric of GPT-4o scoring, Doubao-V2 and Gemini2.5-pro achieve the optimal performance. Qwen2.5-VL-72B-Instruct also achieves a BERTScore close to that of GPT-4o, proving that its training process includes a large amount of ancient text-related data, endowing it with relatively rich knowledge of ancient Chinese. Notably, in knowledge-based QA, Qwen2.5-VL-7B does not demonstrate the same outstanding capabilities as it shows in page-level OCR. This phenomenon suggests that the task imposes higher requirements on the breadth of knowledge possessed by the model, while small and medium-sized models often struggle to cover the knowledge required in the field of history and culture due to the limitations of their pre-training corpus.



\begin{table}[t]
  \centering
  \scalebox{0.8}{
  \begin{tabular}{clcc}
    \toprule
    & \textbf{Model} & \textbf{GPT-4o} & \textbf{BS-F1} \\
    \midrule

    \multirow{14}{*}{\parbox{2cm}{Open-source MLLMs}} & \multicolumn{3}{c}{\textbf{2B-4B MLLMs}} \\
    \cmidrule(lr){2-4}
    & InternVL2.5-2B & 2.25 & \underline{62.24} \\
    & InternVL3-2B & 2.83 & 58.95 \\
    & Qwen2.5-VL-3B & 3.75 & 52.30 \\
    & Qwen2.5-Omni-3B & 3.40 & 56.87 \\
    & InternVL2.5-4B & 3.26 & 58.22 \\

    \addlinespace
    \cmidrule(lr){2-4}
    & \multicolumn{3}{c}{\textbf{7B-8B MLLMs}} \\
    \cmidrule(lr){2-4}
    & Qwen2.5-VL-7B & 4.75 & 57.48 \\
    & Qwen2.5-Omni-7B & 4.12 & 58.62 \\
    & LLaVA-1.5-7B & 0.87 & 56.56 \\
    & InternVL2.5-8B & 3.63 & 61.52 \\
    & InternVL3-8B & 3.53 & 55.96 \\

    \addlinespace
    \cmidrule(lr){2-4}
    & \multicolumn{3}{c}{\textbf{70B+ MLLMs}} \\
    \cmidrule(lr){2-4}
    & Qwen2.5-VL-72B & 5.63 & 59.34 \\
    & LLaVA-OneVision-72B & 3.43 & 57.57 \\
    & InternVL3-78B & 4.13 & 57.78 \\

    \midrule
    \multirow{4}{*}{\parbox{2cm}{Closed-source MLLMs}} & Qwen-VL-Max & 5.67 & 58.77 \\
    & Doubao-V2 & \underline{5.79} & 57.70 \\
    & Gemini2.5-Pro & \textbf{5.92} & 62.06 \\
    & GPT-4o & 5.03 & \textbf{64.58} \\
    \bottomrule
  \end{tabular}}
    \caption{Evaluation on linguistic variant QA of AncientDoc.}
  \label{language}

\end{table}

\noindent\textbf{Results in Linguistic Variant QA.} From the evaluation results in Tab.~\ref{language}, GPT-4o and Gemini 2.5-pro once again demonstrate their leading advantages, achieving BERTScore of 64.58 and 62.06 respectively. The results indicate the reliability of their capabilities in language style transformation and understanding. A notable phenomenon is that the InternVL2.5 series perform significantly better in this task than in previous tasks. Among them, InternVL2.5-2B, with a BERTScore exceeding that of Gemini 2.5-pro, becomes the strongest performer among open-source models. More remarkably, the overall score of the InternVL2.5 series is much higher than that of the InternVL3 series. This contrast may stem from the fact that more ancient-related knowledge and data are introduced in its training process, enabling it to have better modeling capabilities in understanding classical allusions and stylistic features.

\subsection{Comparison between BERTScore and GPT-4o Scoring}

Through the results in Tab.~\ref{ocr}-~\ref{language}, BERTScore ratings largely align with those of GPT-4o scores, showing a positive correlation. This result demonstrates the rationality of selecting GPT-4o as a scoring tool for large models, which also validates our conclusion in Sec.~\ref{Metric}. However, due to the existence of numerous possible outcomes in vernacular translation results, GPT-4o's scoring tends to be conservative. The top-performing model only achieves a score of 4.72 (out of 10) in GPT-4o's evaluation. Even more notably, when evaluating GPT-4o's own performance in vernacular translation, the score is merely 0.92, indicating that it does not exhibit biased scoring towards its own predictions.


\section{Conclusion}

In conclusion, this work addresses the critical gap in Chinese ancient document understanding. We construct the first systematic benchmark called AncientDoc, featuring a multi-task design (page-level OCR, vernacular translation, reasoning-based QA, knowledge-based QA, and linguistic variant QA) with diverse sources (14 types of ancient documents from over 100 books, about 3000 page images) and tailored metrics. Comprehensive metrics, along with using human-consistent LLMs GPT-4o as a metric, are adopted. This benchmark paves the way for exploring ancient Chinese cultural heritage via VLMs, stimulates ancient Chinese research. 


\bibliography{aaai2026}

\begin{thebibliography}{39}
\providecommand{\natexlab}[1]{#1}

\bibitem[{Akoushideh, Ranjkesh~Rashtehroudi, and Shahbahrami(2025)}]{akoushideh2025persian}
Akoushideh, A.; Ranjkesh~Rashtehroudi, A.; and Shahbahrami, A. 2025.
\newblock Persian/Arabic Scene Text Recognition With Convolutional Recurrent Neural Network.
\newblock \emph{IET Smart Cities}, 7(1): e70001.

\bibitem[{Bai et~al.(2023)Bai, Bai, Yang, Wang, Tan, Wang, Lin, Zhou, and Zhou}]{Qwen-VL}
Bai, J.; Bai, S.; Yang, S.; Wang, S.; Tan, S.; Wang, P.; Lin, J.; Zhou, C.; and Zhou, J. 2023.
\newblock Qwen-VL: A Versatile Vision-Language Model for Understanding, Localization, Text Reading, and Beyond.
\newblock \emph{arXiv preprint arXiv:2308.12966}.

\bibitem[{Bai et~al.(2025)Bai, Chen, Liu, Wang, Ge, Song, Dang, Wang, Wang, Tang et~al.}]{bai2025qwen2}
Bai, S.; Chen, K.; Liu, X.; Wang, J.; Ge, W.; Song, S.; Dang, K.; Wang, P.; Wang, S.; Tang, J.; et~al. 2025.
\newblock Qwen2. 5-vl technical report.
\newblock \emph{arXiv preprint arXiv:2502.13923}.

\bibitem[{Borchmann et~al.(2021)Borchmann, Pietruszka, Stanislawek, Jurkiewicz, Turski, Szyndler, and Grali{\'n}ski}]{borchmann2021due}
Borchmann, {\L}.; Pietruszka, M.; Stanislawek, T.; Jurkiewicz, D.; Turski, M.; Szyndler, K.; and Grali{\'n}ski, F. 2021.
\newblock Due: End-to-end document understanding benchmark.
\newblock In \emph{Thirty-fifth Conference on Neural Information Processing Systems Datasets and Benchmarks Track (Round 2)}.

\bibitem[{Chen et~al.(2024)Chen, Wang, Cao, Liu, Gao, Cui, Zhu, Ye, Tian, Liu et~al.}]{chen2024expanding}
Chen, Z.; Wang, W.; Cao, Y.; Liu, Y.; Gao, Z.; Cui, E.; Zhu, J.; Ye, S.; Tian, H.; Liu, Z.; et~al. 2024.
\newblock Expanding performance boundaries of open-source multimodal models with model, data, and test-time scaling.
\newblock \emph{arXiv preprint arXiv:2412.05271}.

\bibitem[{Comanici et~al.(2025)Comanici, Bieber, Schaekermann, Pasupat, Sachdeva, Dhillon, Blistein, Ram, Zhang, Rosen et~al.}]{comanici2025gemini}
Comanici, G.; Bieber, E.; Schaekermann, M.; Pasupat, I.; Sachdeva, N.; Dhillon, I.; Blistein, M.; Ram, O.; Zhang, D.; Rosen, E.; et~al. 2025.
\newblock Gemini 2.5: Pushing the frontier with advanced reasoning, multimodality, long context, and next generation agentic capabilities.
\newblock \emph{arXiv preprint arXiv:2507.06261}.

\bibitem[{Cui, Zhu, and Ding(2025)}]{cui2025enhanced}
Cui, R.; Zhu, A.; and Ding, Z. 2025.
\newblock Enhanced Chinese scene text recognition model base on cross-domain feature fusion.
\newblock \emph{Signal, Image and Video Processing}, 19(6): 499.

\bibitem[{Ding et~al.(2023)Ding, Luo, Chung, and Han}]{ding2023vqa}
Ding, Y.; Luo, S.; Chung, H.; and Han, S.~C. 2023.
\newblock VQA: A new dataset for real-world VQA on PDF documents.
\newblock In \emph{Joint European Conference on Machine Learning and Knowledge Discovery in Databases}, 585--601. Springer.

\bibitem[{Ding et~al.(2024)Ding, Ren, Huang, Luo, and Han}]{ding2024mvqa}
Ding, Y.; Ren, K.; Huang, J.; Luo, S.; and Han, S.~C. 2024.
\newblock MVQA: A dataset for multimodal information retrieval in PDF-based visual question answering.
\newblock \emph{arXiv preprint arXiv:2404.12720}.

\bibitem[{Fu et~al.(2024)Fu, Kuang, Song, Huang, Yang, Li, Zhu, Luo, Wang, Lu et~al.}]{fu2024ocrbench}
Fu, L.; Kuang, Z.; Song, J.; Huang, M.; Yang, B.; Li, Y.; Zhu, L.; Luo, Q.; Wang, X.; Lu, H.; et~al. 2024.
\newblock Ocrbench v2: An improved benchmark for evaluating large multimodal models on visual text localization and reasoning.
\newblock \emph{arXiv preprint arXiv:2501.00321}.

\bibitem[{He et~al.(2016)He, Huang, Qiao, Loy, and Tang}]{he2016reading}
He, P.; Huang, W.; Qiao, Y.; Loy, C.; and Tang, X. 2016.
\newblock Reading scene text in deep convolutional sequences.
\newblock In \emph{Proceedings of the AAAI conference on artificial intelligence}, volume~30.

\bibitem[{Huang et~al.(2022)Huang, Lv, Cui, Lu, and Wei}]{huang2022layoutlmv3}
Huang, Y.; Lv, T.; Cui, L.; Lu, Y.; and Wei, F. 2022.
\newblock Layoutlmv3: Pre-training for document ai with unified text and image masking.
\newblock In \emph{Proceedings of the 30th ACM international conference on multimedia}, 4083--4091.

\bibitem[{Hurst et~al.(2024)Hurst, Lerer, Goucher, Perelman, Ramesh, Clark, Ostrow, Welihinda, Hayes, Radford et~al.}]{hurst2024gpt}
Hurst, A.; Lerer, A.; Goucher, A.~P.; Perelman, A.; Ramesh, A.; Clark, A.; Ostrow, A.; Welihinda, A.; Hayes, A.; Radford, A.; et~al. 2024.
\newblock Gpt-4o system card.
\newblock \emph{arXiv preprint arXiv:2410.21276}.

\bibitem[{Ingle et~al.(2019)Ingle, Fujii, Deselaers, Baccash, and Popat}]{ingle2019scalable}
Ingle, R.~R.; Fujii, Y.; Deselaers, T.; Baccash, J.; and Popat, A.~C. 2019.
\newblock A scalable handwritten text recognition system.
\newblock In \emph{2019 International conference on document analysis and recognition (ICDAR)}, 17--24. IEEE.

\bibitem[{Jaume, Ekenel, and Thiran(2019)}]{jaume2019funsd}
Jaume, G.; Ekenel, H.~K.; and Thiran, J.-P. 2019.
\newblock Funsd: A dataset for form understanding in noisy scanned documents.
\newblock In \emph{2019 International Conference on Document Analysis and Recognition Workshops (ICDARW)}, volume~2, 1--6. IEEE.

\bibitem[{Kang et~al.(2022)Kang, Riba, Rusi{\~n}ol, Forn{\'e}s, and Villegas}]{kang2022pay}
Kang, L.; Riba, P.; Rusi{\~n}ol, M.; Forn{\'e}s, A.; and Villegas, M. 2022.
\newblock Pay attention to what you read: non-recurrent handwritten text-line recognition.
\newblock \emph{Pattern Recognition}, 129: 108766.

\bibitem[{Kang et~al.(2024)Kang, Tito, Valveny, and Karatzas}]{kang2024multi}
Kang, L.; Tito, R.; Valveny, E.; and Karatzas, D. 2024.
\newblock Multi-page document visual question answering using self-attention scoring mechanism.
\newblock In \emph{International Conference on Document Analysis and Recognition}, 219--232. Springer.

\bibitem[{Li et~al.(2024)Li, Zhang, Guo, Zhang, Li, Zhang, Zhang, Zhang, Li, Liu et~al.}]{li2024llava}
Li, B.; Zhang, Y.; Guo, D.; Zhang, R.; Li, F.; Zhang, H.; Zhang, K.; Zhang, P.; Li, Y.; Liu, Z.; et~al. 2024.
\newblock Llava-onevision: Easy visual task transfer.
\newblock \emph{arXiv preprint arXiv:2408.03326}.

\bibitem[{Liu et~al.(2024{\natexlab{a}})Liu, Li, Li, and Lee}]{liu2024improved}
Liu, H.; Li, C.; Li, Y.; and Lee, Y.~J. 2024{\natexlab{a}}.
\newblock Improved baselines with visual instruction tuning.
\newblock In \emph{Proceedings of the IEEE/CVF conference on computer vision and pattern recognition}, 26296--26306.

\bibitem[{Liu et~al.(2024{\natexlab{b}})Liu, Li, Huang, Yang, Yu, Li, Yin, Liu, Jin, and Bai}]{liu2024ocrbench}
Liu, Y.; Li, Z.; Huang, M.; Yang, B.; Yu, W.; Li, C.; Yin, X.-C.; Liu, C.-L.; Jin, L.; and Bai, X. 2024{\natexlab{b}}.
\newblock Ocrbench: on the hidden mystery of ocr in large multimodal models.
\newblock \emph{Science China Information Sciences}, 67(12): 220102.

\bibitem[{Ma et~al.(2024)Ma, Zang, Chen, Chen, Jiao, Li, Lu, Liu, Ma, Dong et~al.}]{ma2024mmlongbench}
Ma, Y.; Zang, Y.; Chen, L.; Chen, M.; Jiao, Y.; Li, X.; Lu, X.; Liu, Z.; Ma, Y.; Dong, X.; et~al. 2024.
\newblock Mmlongbench-doc: Benchmarking long-context document understanding with visualizations.
\newblock \emph{Advances in Neural Information Processing Systems}, 37: 95963--96010.

\bibitem[{Mathew, Karatzas, and Jawahar(2021)}]{mathew2021docvqa}
Mathew, M.; Karatzas, D.; and Jawahar, C. 2021.
\newblock Docvqa: A dataset for vqa on document images.
\newblock In \emph{Proceedings of the IEEE/CVF winter conference on applications of computer vision}, 2200--2209.

\bibitem[{Mishra et~al.(2019)Mishra, Shekhar, Singh, and Chakraborty}]{mishra2019ocr}
Mishra, A.; Shekhar, S.; Singh, A.~K.; and Chakraborty, A. 2019.
\newblock Ocr-vqa: Visual question answering by reading text in images.
\newblock In \emph{2019 international conference on document analysis and recognition (ICDAR)}, 947--952. IEEE.

\bibitem[{Popovi{\'c}(2017)}]{popovic2017chrf++}
Popovi{\'c}, M. 2017.
\newblock chrF++: words helping character n-grams.
\newblock In \emph{Proceedings of the second conference on machine translation}, 612--618.

\bibitem[{Qwen et~al.(2024)Qwen, :, Yang, Yang, Zhang, Hui, Zheng, Yu, Li, Liu, Huang, Wei, Lin, Yang, Tu, Zhang, Yang, Yang, Zhou, Lin, Dang, Lu, Bao, Yang, Yu, Li, Xue, Zhang, Zhu, Men, Lin, Li, Tang, Xia, Ren, Ren, Fan, Su, Zhang, Wan, Liu, Cui, Zhang, and Qiu}]{qwen2024qwen25}
Qwen; :; Yang, A.; Yang, B.; Zhang, B.; Hui, B.; Zheng, B.; Yu, B.; Li, C.; Liu, D.; Huang, F.; Wei, H.; Lin, H.; Yang, J.; Tu, J.; Zhang, J.; Yang, J.; Yang, J.; Zhou, J.; Lin, J.; Dang, K.; Lu, K.; Bao, K.; Yang, K.; Yu, L.; Li, M.; Xue, M.; Zhang, P.; Zhu, Q.; Men, R.; Lin, R.; Li, T.; Tang, T.; Xia, T.; Ren, X.; Ren, X.; Fan, Y.; Su, Y.; Zhang, Y.; Wan, Y.; Liu, Y.; Cui, Z.; Zhang, Z.; and Qiu, Z. 2024.
\newblock Qwen2.5 Technical Report.
\newblock arXiv:2412.15115.

\bibitem[{Tanaka et~al.(2024)Tanaka, Iki, Nishida, Saito, and Suzuki}]{tanaka2024instructdoc}
Tanaka, R.; Iki, T.; Nishida, K.; Saito, K.; and Suzuki, J. 2024.
\newblock Instructdoc: A dataset for zero-shot generalization of visual document understanding with instructions.
\newblock In \emph{Proceedings of the AAAI conference on artificial intelligence}, volume~38, 19071--19079.

\bibitem[{Team(2025)}]{seed2025seed1_5vl}
Team, B.~S. 2025.
\newblock Seed1.5-VL Technical Report.
\newblock \emph{arXiv preprint arXiv:2505.07062}.

\bibitem[{Tian et~al.(2016)Tian, Huang, He, He, and Qiao}]{tian2016detecting}
Tian, Z.; Huang, W.; He, T.; He, P.; and Qiao, Y. 2016.
\newblock Detecting text in natural image with connectionist text proposal network.
\newblock In \emph{European conference on computer vision}, 56--72. Springer.

\bibitem[{Wang et~al.(2023)Wang, Lyu, Ji, Zhang, Yu, Shi, and Tu}]{wang2023document}
Wang, L.; Lyu, C.; Ji, T.; Zhang, Z.; Yu, D.; Shi, S.; and Tu, Z. 2023.
\newblock Document-level machine translation with large language models.
\newblock \emph{arXiv preprint arXiv:2304.02210}.

\bibitem[{Wu et~al.(2024)Wu, Chen, Pan, Liu, Liu, Dai, Gao, Ma, Wu, Wang et~al.}]{wu2024deepseek}
Wu, Z.; Chen, X.; Pan, Z.; Liu, X.; Liu, W.; Dai, D.; Gao, H.; Ma, Y.; Wu, C.; Wang, B.; et~al. 2024.
\newblock Deepseek-vl2: Mixture-of-experts vision-language models for advanced multimodal understanding.
\newblock \emph{arXiv preprint arXiv:2412.10302}.

\bibitem[{Xu et~al.(2025)Xu, Guo, He, Hu, He, Bai, Chen, Wang, Fan, Dang et~al.}]{xu2025qwen2}
Xu, J.; Guo, Z.; He, J.; Hu, H.; He, T.; Bai, S.; Chen, K.; Wang, J.; Fan, Y.; Dang, K.; et~al. 2025.
\newblock Qwen2. 5-omni technical report.
\newblock \emph{arXiv preprint arXiv:2503.20215}.

\bibitem[{Xu and Xiang(2025)}]{xu2025improving}
Xu, Z.; and Xiang, Y. 2025.
\newblock Improving Chinese word segmentation with character--lexicon class attention.
\newblock \emph{Neural Computing and Applications}, 37(5): 3857--3867.

\bibitem[{Yang et~al.(2018)Yang, Jin, Huang, Yang, Lai, and Sun}]{yang2018dense}
Yang, H.; Jin, L.; Huang, W.; Yang, Z.; Lai, S.; and Sun, J. 2018.
\newblock Dense and tight detection of Chinese characters in historical documents: Datasets and a recognition guided detector.
\newblock \emph{IEEE Access}, 6: 30174--30183.

\bibitem[{Yin et~al.(2017)Yin, Wu, Zhang, and Liu}]{yin2017scene}
Yin, F.; Wu, Y.-C.; Zhang, X.-Y.; and Liu, C.-L. 2017.
\newblock Scene text recognition with sliding convolutional character models.
\newblock \emph{arXiv preprint arXiv:1709.01727}.

\bibitem[{Zhang et~al.(2018)Zhang, Luan, Sun, Zhai, Xu, Zhang, and Liu}]{zhang2018improving}
Zhang, J.; Luan, H.; Sun, M.; Zhai, F.; Xu, J.; Zhang, M.; and Liu, Y. 2018.
\newblock Improving the transformer translation model with document-level context.
\newblock \emph{arXiv preprint arXiv:1810.03581}.

\bibitem[{Zhang, Yu, and Zhang(2024)}]{zhang2024cream}
Zhang, J.; Yu, Y.; and Zhang, Y. 2024.
\newblock CREAM: coarse-to-fine retrieval and multi-modal efficient tuning for document VQA.
\newblock In \emph{Proceedings of the 32nd ACM International Conference on Multimedia}, 925--934.

\bibitem[{Zhang et~al.(2019)Zhang, Kishore, Wu, Weinberger, and Artzi}]{zhang2019bertscore}
Zhang, T.; Kishore, V.; Wu, F.; Weinberger, K.~Q.; and Artzi, Y. 2019.
\newblock Bertscore: Evaluating text generation with bert.
\newblock \emph{arXiv preprint arXiv:1904.09675}.

\bibitem[{Zhou et~al.(2017)Zhou, Yao, Wen, Wang, Zhou, He, and Liang}]{zhou2017east}
Zhou, X.; Yao, C.; Wen, H.; Wang, Y.; Zhou, S.; He, W.; and Liang, J. 2017.
\newblock East: an efficient and accurate scene text detector.
\newblock In \emph{Proceedings of the IEEE conference on Computer Vision and Pattern Recognition}, 5551--5560.

\bibitem[{Zhu et~al.(2025)Zhu, Wang, Chen, Liu, Ye, Gu, Tian, Duan, Su, Shao et~al.}]{zhu2025internvl3}
Zhu, J.; Wang, W.; Chen, Z.; Liu, Z.; Ye, S.; Gu, L.; Tian, H.; Duan, Y.; Su, W.; Shao, J.; et~al. 2025.
\newblock Internvl3: Exploring advanced training and test-time recipes for open-source multimodal models.
\newblock \emph{arXiv preprint arXiv:2504.10479}.

\end{thebibliography}



\appendix
\section{Explanations of 14 Document Categories in AncientDoc}

The 14 categories of Chinese ancient documents included in AncientDoc cover a wide range of genres, reflecting the diversity of knowledge and literary forms in traditional Chinese culture. Their specific explanations are as follows:  
\begin{itemize}
    \item Biography: These texts focus on recording the life stories, achievements, and moral virtues of historical figures, typically presenting chronological narratives that serve as moral examples for readers. 
    \item Confucianism: Works centered on Confucian philosophy and ethics, including interpretations of the "Five Classics" (e.g., Book of Changes, Book of Songs) and writings by Confucian scholars that elaborate on concepts like benevolence, righteousness, and propriety.  
    \item The Philosophy of War: Military strategy texts, such as Sun Tzu’s Art of War, which explore tactics, command principles, and the philosophy of warfare, emphasizing strategic thinking and battlefield wisdom. 
    \item Collected Works: Comprehensive anthologies compiling poems, essays, or other writings by multiple authors (or a single author across genres), often organized thematically (e.g., nature, politics) or chronologically.  
    \item Medical Authors: Ancient medical texts that discuss diagnostic methods, herbal remedies, acupuncture techniques, and classical medical theories (e.g., yin-yang and five elements), forming the foundation of traditional Chinese medicine.  
    \item Astronomy and Mathematics: Scientific treatises on calendrical systems, astronomical observations (e.g., eclipses, planetary movements), and mathematical operations, reflecting ancient China’s advancements in natural sciences.  
    \item Masters of Unofficial History: Semi-historical or fictional works that offer alternative perspectives on historical events, anecdotes, or folklore, often filling gaps in official historical records with vivid narratives.
    \item General Anthologies: Anthologies dedicated to the literary works of a single author, distinct from "Collected Works" (which include multiple authors). These texts highlight the unique style and creative characteristics of individual writers. 
    \item Eclectics: Works that integrate ideas from multiple philosophical schools, such as Confucianism, Taoism, and Legalism, aiming to synthesize diverse thoughts into a coherent worldview.
    \item Chuci-style Poetry: Poetry modeled after the Chuci (Songs of Chu), a classic collection of ancient Chinese poetry. This genre is known for its rich imagery, mythological allusions, and emotional intensity, often expressing patriotic or melancholic sentiments. 
    \item Leishu: Traditional encyclopedic works organized by topic (e.g., astronomy, geography, literature), serving as reference tools for scholars and educators to access comprehensive knowledge efficiently.  
    \item Art: Texts focusing on traditional Chinese arts, including painting, calligraphy, music, and handicrafts, discussing techniques, aesthetic principles, and the cultural significance of artistic creation. 
    \item Literary Criticism of Poetry and Prose: Analytical writings that evaluate classical literature, focusing on literary form, rhetorical devices, and artistic merit. These works shape critical standards for poetry, essays, and other genres.  
    \item Genealogies and Catalogues: Documents recording family lineages (genealogies), bibliographic records of texts, or catalogs of artifacts (e.g., antiques, books), playing a key role in preserving historical and cultural heritage.
\end{itemize}

Together, these 14 categories encompass historical records, philosophical treatises, literary works, scientific texts, and practical references, providing a comprehensive sample of Chinese ancient documents for evaluating VLMs’ cross-domain understanding capabilities.

\section{Evaluation Metrics}

To rigorously assess the performance of Vision-Language Models (VLMs) on the tasks in AncientDoc, we employ a set of metrics tailored to different task characteristics, as detailed below:

\subsection{Metrics for Page-level OCR}
Four character-level metrics are used to evaluate the accuracy of text extraction from ancient document pages:

\textbf{Character Error Rate (CER)}: Measures the normalized number of character-level errors (substitutions, insertions, deletions) between the predicted sequence and the reference text:
\begin{equation}
   \mathrm{CER} = \frac{S + I + D}{|T|}
   \end{equation}
   where \( S \), \( I \), and \( D \) represent the counts of substituted, inserted, and deleted characters, respectively, and \( |T| \) is the length of the reference sequence (Ground Truth).

\textbf{Character Precision}: Reflects the proportion of correctly predicted characters relative to the total number of characters in the model's output:
   \begin{equation}
   \mathrm{CharPrecision} = \frac{M}{|P|}
   \end{equation}
   where \( M \) is the number of correctly predicted characters, and \( |P| \) is the total number of characters in the predicted sequence.

\textbf{Character Recall}: Measures the proportion of correctly predicted characters relative to the total number of characters in the reference text:
   \begin{equation}
   \mathrm{CharRecall} = \frac{M}{|T|}
   \end{equation}
   where \( |T| \) denotes the length of the reference sequence.

\textbf{Character F1}: A harmonic mean of Character Precision and Character Recall, balancing both metrics to provide a comprehensive evaluation of OCR accuracy:
   \begin{equation}
   \mathrm{CharF1} = 2 \cdot \frac{\mathrm{CharPrecision} \cdot \mathrm{CharRecall}}{\mathrm{CharPrecision} + \mathrm{CharRecall}}
   \end{equation}

\subsection{Metrics for Higher-Level Understanding Tasks}
For vernacular translation, reasoning-based QA, knowledge-based QA, and linguistic variant QA, the following metrics are adopted:

\textbf{CHRF++}: Computes the F-score based on character and word n-grams, with \( \beta = 2 \) to emphasize recall:
   \begin{equation}
   \mathrm{CHRF^{++}} = F_{\beta}(\text{char-n-gram}, \text{word-n-gram})
   \end{equation}

\textbf{BERTScore (BS-F1)}: A semantic-level metric that calculates precision and recall using cosine similarities between BERT embeddings of predicted and reference texts, then derives the F1 score:
   \begin{equation}
   \mathrm{BS\text{-}F1}(P, T) = \frac{2 \cdot \mathrm{Precision} \cdot \mathrm{Recall}}{\mathrm{Precision} + \mathrm{Recall}}
   \end{equation}

\textbf{GPT-4o Score}: A human-aligned metric where GPT-4o scores model predictions against references on a scale of 0–10. The final score is the average of all individual scores:
   \begin{equation}
   \mathrm{GPT4o\_Score} = \frac{1}{N} \sum_{i=1}^{N} s_i
   \end{equation}
   where \( s_i \) is the score for the \( i \)-th sample, and \( N \) is the total number of samples.

\section{Metrics for Evaluating Differences Between Human and LLM Scoring}

To quantify the consistency and discrepancy between human scoring and large model scoring, we employ six statistical metrics, defined as follows:

Pearson correlation coefficient: Measures the linear correlation between two sets of scores ($x$ for human scores, $y$ for model scores):
   \begin{equation}
   \mathrm{Pearson}(x, y) = \frac{\sum_{i=1}^n (x_i - \bar{x})(y_i - \bar{y})}{\sqrt{\sum_{i=1}^n (x_i - \bar{x})^2} \cdot \sqrt{\sum_{i=1}^n (y_i - \bar{y})^2}} 
   \end{equation}
   where $\bar{x}$ and $\bar{y}$ denote the means of $x$ and $y$, respectively.

Spearman rank correlation coefficient: Evaluates the monotonic relationship by comparing score ranks:
   \begin{equation}
   \mathrm{Spearman}(x, y) = 1 - \frac{6 \sum_{i=1}^n d_i^2}{n(n^2 - 1)} 
   \end{equation}
   with $d_i = \text{rank}(x_i) - \text{rank}(y_i)$ representing the rank difference of the $i$-th sample, and $n$ being the total number of samples.

Kendall tau coefficient: Assesses ordinal association by counting concordant and discordant pairs:
   \begin{equation}
   \mathrm{Kendall}(x, y) = \frac{C - D}{\frac{1}{2}n(n - 1)} 
   \end{equation}
   where $C$ is the number of concordant pairs (consistent order) and $D$ is the number of discordant pairs (inconsistent order).

Mean Squared Error (MSE): Quantifies the average squared difference between scores:
   \begin{equation}
   \mathrm{MSE}(x, y) = \frac{1}{n} \sum_{i=1}^{n} (x_i - y_i)^2 
   \end{equation}

Mean Absolute Error (MAE): Measures the average absolute difference between scores:
   \begin{equation}
   \mathrm{MAE}(x, y) = \frac{1}{n} \sum_{i=1}^{n} |x_i - y_i| 
   \end{equation}

Bias: Represents the systematic deviation between the mean of human scores and model scores:
   \begin{equation}
   \mathrm{Bias}(x, y) = \frac{1}{n} \sum_{i=1}^{n} x_i - \frac{1}{n} \sum_{i=1}^{n} y_i = \bar{x} - \bar{y} 
   \end{equation}

These metrics collectively capture linear/correlation trends, ordinal relationships, error magnitudes, and systematic deviations, providing a comprehensive assessment of scoring consistency.

\section{Consistency Analysis Between Human Scoring and LLM Scoring Across Five Subtasks}

\begin{figure*}[h]
\centering
\includegraphics[width=1.0\textwidth]{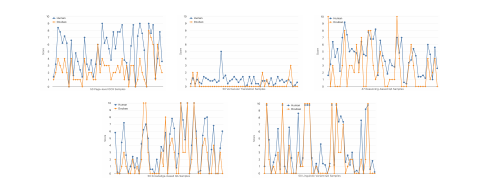} 
\caption{Comparison of consistency between Doubao scoring and human scoring.}
\label{Doubao}
\end{figure*}

\begin{figure*}[h]
\centering
\includegraphics[width=1.0\textwidth]{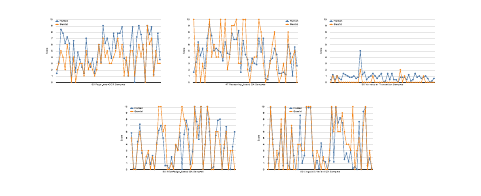} 
\caption{Comparison of consistency between Gemini scoring and human scoring.}
\label{Gemini}
\end{figure*}

\begin{figure*}[h]
\centering
\includegraphics[width=1.0\textwidth]{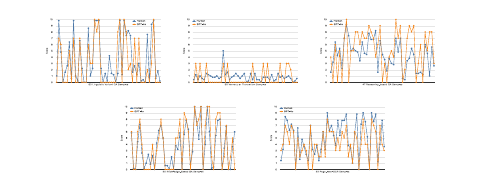} 
\caption{Comparison of consistency between GPT-4o scoring and human scoring.}
\label{GPT-4o}
\end{figure*}

\begin{figure*}[h]
\centering
\includegraphics[width=1.0\textwidth]{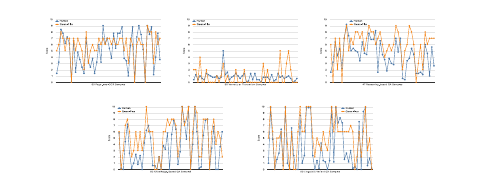} 
\caption{Comparison of consistency between Qwen-Plus scoring and human scoring.}
\label{qwenplus}
\end{figure*}

\begin{figure*}[h]
\centering
\includegraphics[width=1.0\textwidth]{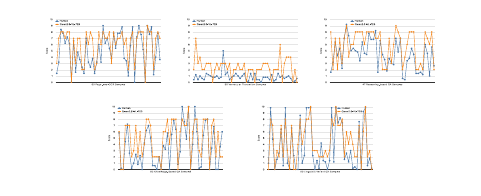} 
\caption{Comparison of consistency between Qwen-VL-72B scoring and human scoring.}
\label{qwenvl72}
\end{figure*}

To evaluate the consistency between human scoring and large model scoring across the five subtasks of AncientDoc, we compare the scoring results of multiple mainstream large models (including Qwen2.5-VL-72B, Gemini, Doubao, Qwen-Plus, and GPT-4o) with human ratings using six metrics: Pearson, Spearman, Kendall, MSE, MAE, and Bias. The results indicate that GPT-4o exhibits the highest alignment with human scoring across all subtasks (as shown in Fig.~\ref{Doubao}-~\ref{qwenvl72}).

\section{All evaluation results}

Since the limited length of the main text, we only show some results in the main text. In the following table, we have displayed all evaluation results (shown in Tab.~\ref{tab_ocr}-~\ref{tab_language}).

\begin{table*}[h!]
\centering
\caption{Evaluation Results on Page-level OCR of AncientDoc}
\label{tab_ocr}
\begin{tabular}{lcccccc}
\toprule
\textbf{Model} & \textbf{CER} & \textbf{CharPrecision} & \textbf{CharRecall} & \textbf{CharF1} & \textbf{GPT-4o} \\
\midrule
Qwen-VL-Max & 66.39 & 9.62 & 9.93 & 9.77 & 5.7 \\
DeepSeek-VL2 & 147.28 & 0.25 & 0.22 & 0.11 & 0.6 \\
InternVL2.5-1B & 97.77 & 2.38 & 1.92 & 2.13 & 2.63 \\
InternVL2.5-2B & 120.72 & 1.76 & 1.69 & 1.72 & 2.03 \\
InternVL2.5-4B & 79.75 & 5.27 & 4.05 & 4.58 & 3.65 \\
InternVL2.5-8B & 96.88 & 2.87 & 2.59 & 2.72 & 3 \\
Qwen2-VL-2B & 81.75 & 4 & 2.66 & 3.2 & 3.79 \\
Qwen2-VL-7B & 71.61 & 7.99 & 6.8 & 7.35 & 4.66 \\
Qwen2.5-VL-3B & 50.36 & 9.44 & 9.02 & 9.23 & 5.56 \\
Qwen2.5-VL-7B & 35.47 & 12.95 & 12.75 & 12.85 & 6.37 \\
Qwen2.5-VL-32B & 59.65 & 14.72 & 14.65 & 14.68 & 5.67 \\
Qwen2.5-VL-72B & 58.83 & 10.62 & 9.56 & 10.06 & 5.73 \\
Qwen2.5-Omni-3B & 63.42 & 5.4 & 4.78 & 5.07 & 4.56 \\
Qwen2.5-Omni-7B & 70.46 & 3.44 & 3.35 & 3.39 & 4.3 \\
Doubao-V2 & 71.95 & 13.14 & 20.44 & 16 & 6.27 \\
Gemini2.5-Pro & 32.03 & 17.73 & 18.53 & 18.12 & 6.08 \\
GPT-4o & 75.1 & 4.83 & 2.72 & 3.48 & 2.97 \\
InternVL3-1B & 97.49 & 3.57 & 1.83 & 2.42 & 1.83 \\
InternVL3-2B & 95.59 & 2.46 & 1.79 & 2.07 & 2.52 \\
InternVL3-8B & 51.8 & 7.3 & 6.67 & 6.97 & 5.06 \\
InternVL3-38B & 103.02 & 3.26 & 3.75 & 3.49 & 2.98 \\
InternVL3-78B & 78.67 & 4.7 & 4.89 & 4.79 & 4.04 \\
LLaVA-Onevision-5B & 215.31 & 0.05 & 0.07 & 0.01 & 0 \\
LLaVA-Onevision-72B & 190.67 & 0.29 & 0.43 & 0.25 & 0.01 \\
LLaVA1.5-7B & 129.71 & 0.04 & 0.03 & 0 & 0.05 \\
\bottomrule
\end{tabular}
\end{table*}

\begin{table}[h!]
\centering
\caption{Evaluation Results on Reasoning-based QA of AncientDoc}
\label{tab_reasoning}
\begin{tabular}{lccc}
\toprule
\textbf{Model} & \textbf{CHRF++} & \textbf{BS-F1} & \textbf{GPT-4o} \\
\midrule
Qwen-VL-Max & 8.6 & 71.3 & 7.46 \\
DeepSeek-VL2 & 3.42 & 59.09 & 1.01 \\
InternVL2.5-1B & 4.58 & 63.8 & 2.38 \\
InternVL2.5-2B & 5.01 & 65.29 & 3.55 \\
InternVL2.5-4B & 5.77 & 66.12 & 4.75 \\
InternVL2.5-8B & 7.47 & 68.4 & 5.93 \\
Qwen2-VL-2B & 6.12 & 67.8 & 4.45 \\
Qwen2-VL-7B & 6.91 & 69.03 & 5.79 \\
Qwen2.5-VL-3B & 4.8 & 65.83 & 4.67 \\
Qwen2.5-VL-7B & 7.34 & 69.96 & 6.44 \\
Qwen2.5-VL-32B & 9.69 & 70.9 & 7.49 \\
Qwen2.5-VL-72B & 9.04 & 71.4 & 7.43 \\
Qwen2.5-Omni-3B & 5.63 & 66.96 & 4.9 \\
Qwen2.5-Omni-7B & 6.52 & 68.7 & 6.05 \\
Doubao-V2 & 7.15 & 68.78 & 7.4 \\
Gemini2.5-Pro & 8.68 & 69.33 & 7.76 \\
GPT-4o & 7.99 & 70.52 & 6.92 \\
InternVL3-1B & 5.12 & 64.24 & 2.85 \\
InternVL3-2B & 5.66 & 66.45 & 4.39 \\
InternVL3-8B & 5.21 & 65.62 & 4.98 \\
InternVL3-38B & 5.36 & 67.29 & 5.78 \\
InternVL3-78B & 4.95 & 65.99 & 5.18 \\
LLaVA-Onevision-5B & 2.21 & 55.79 & 0.89 \\
LLaVA-Onevision-72B & 5.96 & 67.2 & 5.48 \\
LLaVA1.5-7B & 3.99 & 61.69 & 1.64 \\
\bottomrule
\end{tabular}
\end{table}

\begin{table}[h!]
\centering
\caption{Evaluation Results on Vernacular Translation of AncientDoc}
\label{tab_trans}
\begin{tabular}{lccc}
\toprule
\textbf{Model} & \textbf{CHRF++} & \textbf{BS-F1} & \textbf{GPT-4o} \\
\midrule
Qwen-VL-Max & 12.3 & 71.03 & 3.22 \\
DeepSeek-VL2 & 0.49 & 50.27 & 0 \\
InternVL2.5-1B & 1.35 & 52.73 & 0.02 \\
InternVL2.5-2B & 1.33 & 53.92 & 0.03 \\
InternVL2.5-4B & 2.84 & 58.46 & 0.53 \\
InternVL2.5-8B & 3.2 & 59.24 & 0.58 \\
Qwen2-VL-2B & 1.9 & 52.31 & 0.46 \\
Qwen2-VL-7B & 4.34 & 60.74 & 1.17 \\
Qwen2.5-VL-3B & 1.63 & 55.2 & 0.66 \\
Qwen2.5-VL-7B & 7.04 & 65.59 & 2.3 \\
Qwen2.5-VL-32B & 9.9 & 67.9 & 2.87 \\
Qwen2.5-VL-72B & 9.77 & 69.87 & 2.98 \\
Qwen2.5-Omni-3B & 2.06 & 56.8 & 0.56 \\
Qwen2.5-Omni-7B & 1.13 & 53.99 & 0.46 \\
Doubao-V2 & 0.56 & 52.39 & 2.12 \\
Gemini2.5-Pro & 11.41 & 72.5 & 4.72 \\
GPT-4o & 3.02 & 58.86 & 0.92 \\
InternVL3-1B & 1.86 & 54.92 & 0.06 \\
InternVL3-2B & 0.97 & 51.83 & 0.1 \\
InternVL3-8B & 0.85 & 53.16 & 0.29 \\
InternVL3-38B & 1.99 & 56.93 & 0.56 \\
InternVL3-78B & 4.45 & 62.4 & 1.21 \\
LLaVA-Onevision-5B & 0.76 & 50.85 & 0 \\
LLaVA-Onevision-72B & 0.44 & 48.83 & 0.03 \\
LLaVA1.5-7B & 0.43 & 50 & 0.01 \\
\bottomrule
\end{tabular}
\end{table}

\begin{table}[h!]
\centering
\caption{Evaluation Results on Knowledge-based QA of AncientDoc}
\label{tab_know}
\begin{tabular}{lccc}
\toprule
\textbf{Model} & \textbf{CHRF++} & \textbf{BS-F1} & \textbf{GPT-4o} \\
\midrule
Qwen-VL-Max & 7.58 & 68.67 & 6.78 \\
DeepSeek-VL2 & 3.77 & 59.83 & 0.94 \\
InternVL2.5-1B & 3.73 & 61.59 & 1.68 \\
InternVL2.5-2B & 4.47 & 64.24 & 2.48 \\
InternVL2.5-4B & 4.95 & 64.81 & 3.92 \\
InternVL2.5-8B & 6.93 & 67.68 & 4.88 \\
Qwen2-VL-2B & 5.27 & 66.6 & 3.37 \\
Qwen2-VL-7B & 5.75 & 66.82 & 4.83 \\
Qwen2.5-VL-3B & 4.47 & 62.86 & 3.77 \\
Qwen2.5-VL-7B & 5.87 & 66.75 & 5.23 \\
Qwen2.5-VL-32B & 9.44 & 69.35 & 6.84 \\
Qwen2.5-VL-72B & 7.82 & 69.15 & 6.84 \\
Qwen2.5-Omni-3B & 5.33 & 66.1 & 4.11 \\
Qwen2.5-Omni-7B & 5.67 & 66.85 & 4.94 \\
Doubao-V2 & 8.75 & 69.15 & 7.36 \\
Gemini2.5-Pro & 8.88 & 68.94 & 7.36 \\
GPT-4o & 8.02 & 70.01 & 6.53 \\
InternVL3-1B & 4.46 & 63.05 & 1.82 \\
InternVL3-2B & 5.43 & 66.21 & 3.52 \\
InternVL3-8B & 4.75 & 63.6 & 4.31 \\
InternVL3-38B & 4.84 & 65.45 & 5.11 \\
InternVL3-78B & 5.25 & 65.79 & 5.18 \\
LLaVA-Onevision-5B & 2.32 & 56.67 & 0.51 \\
LLaVA-Onevision-72B & 5.56 & 66.31 & 5.04 \\
LLaVA1.5-7B & 3.23 & 60.37 & 0.78 \\
\bottomrule
\end{tabular}
\end{table}

\begin{table}[h!]
\centering
\caption{Evaluation Results on Linguistic Variant QA of AncientDoc}
\label{tab_language}
\begin{tabular}{lccc}
\toprule
\textbf{Model} & \textbf{CHRF++} & \textbf{BS-F1} & \textbf{GPT-4o} \\
\midrule
Qwen-VL-Max & 3.31 & 58.77 & 5.67 \\
DeepSeek-VL2-Tiny & 3.7 & 60.05 & 0.91 \\
InternVL2.5-1B & 2.61 & 59.27 & 1.59 \\
InternVL2.5-2B & 3.42 & 62.24 & 2.25 \\
InternVL2.5-4B & 2.85 & 58.22 & 3.26 \\
InternVL2.5-8B & 3.65 & 61.52 & 3.63 \\
Qwen2-VL-2B & 3.49 & 58.97 & 3.1 \\
Qwen2-VL-7B & 2.3 & 56.73 & 3.67 \\
Qwen2.5-VL-3B & 1.03 & 52.3 & 3.75 \\
Qwen2.5-VL-7B & 2.65 & 57.48 & 4.75 \\
Qwen2.5-VL-32B & 4.62 & 61.18 & 5.61 \\
Qwen2.5-VL-72B & 3.65 & 59.34 & 5.63 \\
Qwen2.5-Omni-3B & 2.12 & 56.87 & 3.4 \\
Qwen2.5-Omni-7B & 2.66 & 58.62 & 4.12 \\
Doubao-V2 & 3.32 & 57.7 & 5.79 \\
Gemini2.5-Pro & 5.22 & 62.06 & 5.92 \\
GPT-4o & 4.16 & 64.58 & 5.03 \\
InternVL3-1B & 3.69 & 61.92 & 1.51 \\
InternVL3-2B & 2.9 & 58.95 & 2.83 \\
InternVL3-8B & 1.93 & 55.96 & 3.53 \\
InternVL3-38B & 1.96 & 56.87 & 3.99 \\
InternVL3-78B & 2.38 & 57.78 & 4.13 \\
LLaVA-Onevision-5B & 1.72 & 56.12 & 0.48 \\
LLaVA-Onevision-72B & 2.51 & 57.57 & 3.43 \\
LLaVA1.5-7B & 2.27 & 56.56 & 0.87 \\
\bottomrule
\end{tabular}
\end{table}






\end{document}